\documentclass[10pt]{article}

\usepackage{amssymb}
\usepackage{amsmath}
\usepackage{soul}       
\usepackage{xcolor}     
\usepackage{booktabs}   
\usepackage{graphicx}   
\usepackage{mdframed}   
\usepackage{float}      
\usepackage{geometry}   
\usepackage{mathptmx}   
\usepackage{natbib} 

\title{Enhancing Pneumonia Diagnosis and Severity Assessment through Deep Learning: A Comprehensive Approach Integrating CNN Classification and Infection Segmentation}

\author{
    S Kumar Reddy Mallidi\thanks{Corresponding author: \texttt{satya.cnis@gmail.com}} \\
    \textit{Computer Science and Engineering,} \\
    \textit{Sri Vasavi Engineering College,} \\
    Tadepalligudem, AP 534101, India
}
\date{}  
\begin{document}

\maketitle

\begin{abstract}
Lung disease poses a substantial global health challenge, with pneumonia being a prevalent concern. This research focuses on leveraging deep learning techniques to detect and assess pneumonia, addressing two interconnected objectives. Initially, Convolutional Neural Network (CNN) models are introduced for pneumonia classification, emphasizing the necessity of comprehensive diagnostic assessments considering COVID-19. Subsequently, the study advocates for the utilization of deep learning-based segmentation to determine the severity of infection. This dual-pronged approach offers valuable insights for medical professionals, facilitating a more nuanced understanding and effective treatment of pneumonia. Integrating deep learning aims to elevate the accuracy and efficiency of pneumonia detection, thereby contributing to enhanced healthcare outcomes on a global scale.
\end{abstract}


\section*{Research Highlights}
\begin{itemize}
    \item \textbf{Dual-Model Approach for Comprehensive Diagnosis:} Developed two separate deep-learning models for pneumonia classification and infection segmentation, enabling a detailed understanding of both disease presence and severity from chest X-ray images.
    \item \textbf{Utilization of Transfer Learning for Enhanced Classification:} Leveraged the encoder of a pre-trained UNET model and fine-tuned it for infection segmentation as the feature extractor for the pneumonia classification model, improving classification accuracy by focusing on infection regions.
    \item \textbf{Dense Block Inspired Enhancement:} Enhanced the classification model by integrating a dense block inspired by DenseNet-121 to build on top of the VGG16 encoder architecture, improving feature extraction and classification performance.
    \item \textbf{High Accuracy in COVID-19 and Pneumonia Diagnosis:} Achieved significant classification accuracy and segmentation performance, effectively addressing the real-world challenges of limited medical data and variability in X-ray imaging quality.
\end{itemize}

\section{Introduction}
The COVID-19 pandemic has underscored the critical need for accurate and efficient diagnostic tools, especially in medical imaging. COVID-19 manifests across a wide range of severities, from asymptomatic cases to severe respiratory complications, making timely and precise diagnosis essential for effective management and containment. This study leverages deep learning techniques to advance the diagnostic capabilities of chest X-ray imaging, particularly focusing on the intricate features of COVID-19 pathology \cite{Wang2019}.

Deep learning, especially convolutional neural networks (CNNs), has transformed medical image analysis. CNNs have demonstrated notable success in handling high-resolution and large-scale medical imaging datasets, such as the 'ChestX-ray8' database, which includes over 100,000 images. These networks have been pivotal in identifying and localizing pathologies in large image datasets where traditional methods struggle \cite{Wang2019}. CNNs' ability to automatically extract features from raw images provides a streamlined and efficient approach to disease identification in medical imaging, making them highly suited for complex tasks.

Though transformers have emerged in medical imaging due to their self-attention mechanisms, their high computational requirements and need for large datasets present limitations. For instance, transformers require significant training data, which can be challenging to obtain in medical contexts due to privacy concerns and limited labeled datasets \cite{Shamshad2022}. CNNs, on the other hand, have shown superior results in smaller datasets and have established their applicability in clinical settings where computational efficiency is critical.

Integrating transfer learning in CNNs further enhances their effectiveness by leveraging pre-trained models, reducing data and computational demands, and facilitating rapid model adaptation to new tasks \cite{Salehi2023}. Transfer learning has allowed CNNs to perform remarkably well in cases with limited data, proving advantageous in real-world applications such as pneumonia detection, where training data may be scarce.

Moreover, CNNs offer a robust framework for segmentation and classification in medical imaging, with notable adaptability to varying imaging conditions and pathology presentations \cite{Tong2018}. This adaptability allows CNNs to be readily incorporated into clinical workflows, where accurate segmentation and localization of infections are essential for treatment planning and prognosis.

\subsection{Literature Review}
The field of deep learning in medical imaging has grown rapidly, with numerous studies exploring CNN's efficacy in detecting and classifying diseases from X-ray images. For instance, \cite{Sharma2023} reports on the successful application of CNNs in pneumonia detection, demonstrating superior accuracy over traditional methods. Similarly, the CheXImageNet model introduced in \cite{Shastri2022} achieved 100\% accuracy in classifying COVID-19 and other respiratory diseases from chest X-rays, underscoring CNNs' diagnostic capabilities.

The role of CNNs in segmenting lung infections, particularly in COVID-19, has also been highlighted in \cite{Negi2020}. Here, a hybrid model combining GANs with CNNs provided high segmentation accuracy, addressing the challenge of distinguishing infection boundaries in complex X-ray images. Another innovative approach to CNN-based segmentation is seen in \cite{Alom2018}, where recurrent convolutional architectures (RU-Net and R2U-Net) enhance segmentation precision, particularly for medical images with subtle pathological changes.

In addition to CNNs, transfer learning has played a critical role in medical imaging applications. By leveraging pre-trained networks, models can be adapted to new datasets with minimal retraining. The benefits of this approach are discussed in \cite{Salehi2023}, where transfer learning enhances CNN performance in classifying brain and lung diseases, even with limited labeled data. This method's efficiency in handling small datasets is further supported by findings in \cite{Manickam2021}, where transfer learning was key in enhancing CNN-based pneumonia detection.

While transformers have demonstrated potential in medical imaging, as outlined by \cite{Shamshad2022}, high computational demands and a requirement for large datasets often limit their application. Studies such as \cite{Singh2024} highlight that while transformers can capture spatial dependencies across large imaging datasets, CNNs often outperform them in scenarios with constrained computational resources and smaller datasets.

Research comparing CNNs and transformers, such as \cite{Gai2024}, further supports CNNs' advantages in handling high-resolution images with minimal data, where transformers struggle. Additionally, the study by \cite{Rajpurkar2017} on the CheXNet model demonstrates CNNs' capability in achieving state-of-the-art accuracy in pneumonia detection, showing their adaptability and high accuracy.

The application of CNNs in COVID-19 diagnostics is well-supported by \cite{Fan2020}, where the Inf-Net model was developed for COVID-19 infection segmentation, outperforming existing models by a considerable margin. Similarly, the segmentation model in \cite{Afshar2021}, which utilizes CNNs for slice-level classifications, demonstrates CNNs' robust segmentation capabilities, which are vital for accurate infection mapping in clinical diagnostics.

While transformers offer a promising direction in medical imaging, especially for large and complex datasets, CNNs remain advantageous due to their computational efficiency, lower data requirements, and established performance in medical imaging tasks. Studies such as \cite{Shamshad2022} affirm that CNNs, especially when combined with transfer learning, provide a viable and efficient alternative to transformers in medical image analysis, particularly for applications requiring segmentation and classification.

\subsection{Contributions}

This study presents several key contributions to medical image analysis by applying CNN-based models and transfer learning for accurate pneumonia and COVID-19 diagnosis. Our primary contributions include:

\begin{itemize}
    \item \textbf{Development of Two Specialized Models for Pneumonia Classification and Infection Segmentation:} We developed two distinct CNN models, one specifically for pneumonia classification and another for infection segmentation. The segmentation model was trained using a specialized dataset to accurately localize chest X-rays' infection regions. The encoder from the segmentation model was subsequently fine-tuned to enhance the classification model's performance, concentrating particularly on the infection zones.

    \item \textbf{Integration of a Dense Block for Enhanced Classification:} Inspired by the DenseNet121 architecture used in the base of CheXNet, the classification model was built by adding a dense block on top of the four VGG16 convolution blocks extracted from the encoder of the U-Net used for infection segmentation. This combination effectively leverages dense connectivity for better feature reuse and the robust features extracted by the VGG16, improving pneumonia and COVID-19 classification accuracy.

    \item \textbf{Integration of Transfer Learning for Enhanced Performance with Limited Data:} To overcome the data scarcity often present in medical imaging, our study integrates transfer learning to adapt pre-trained CNN models for COVID-19 and pneumonia detection. This approach optimizes performance while reducing the dependency on large, annotated datasets, aligning with findings in \cite{Salehi2023, Manickam2021} and demonstrating the adaptability of CNNs in low-data environments.

    \item \textbf{Gradual Fine-Tuning Strategy for Improved Generalization:} A multi-stage fine-tuning approach was implemented, training only the custom top layers and progressively unfreezing additional layers of the pre-trained VGG16 model. This gradual unfreezing allowed for efficient learning while mitigating overfitting, achieving better generalization even in a limited data setting.

    \item \textbf{Focus on Model Interpretability:} Grad-CAM visualizations enabled enhanced prediction transparency, providing clinicians with a better understanding of the model's decision-making process, as highlighted in \cite{Wang2019, Tong2018}.

\end{itemize}

\section{Related Work}

The related work encompasses a breadth of research efforts to leverage deep learning techniques for COVID-19 diagnosis, pneumonia detection, and semantic segmentation in medical imaging. Wang et al. \cite{Wang2019} have explored deep learning models for COVID-19 diagnosis using chest X-ray and CT images, showcasing promising accuracy in distinguishing COVID-19 cases from other pneumonia types. Semantic segmentation techniques, notably exemplified by the U-Net architecture as introduced by Ronneberger et al. \cite{Ronneberger2015}, have been employed by Salehi et al. \cite{Salehi2023} and Çiçek et al. (2016) to segment lung regions and identify abnormalities in chest X-ray and CT images. Additionally, the importance of explainable deep learning in healthcare, exemplified by attention mechanisms and Grad-CAM visualization as discussed by Rajpurkar et al. \cite{Rajpurkar2017} and Selvaraju et al. \cite{Selvaraju2017}, has been emphasized to aid clinicians in understanding and validating AI-driven diagnostic outcomes. Pneumonia detection studies, including CheXNet by Rajpurkar et al. \cite{Rajpurkar2017}, have demonstrated the feasibility of deep learning models in accurately identifying pneumonia on chest X-rays. Finally, rigorous validation efforts, such as those conducted by Wang et al. \cite{Wang2019} with the COVID-Net initiative and Wang et al. \cite{Wang2017} with the ChestX-ray8 dataset, have provided benchmark datasets and evaluation frameworks for assessing the performance of deep learning models in COVID-19 diagnosis and pneumonia detection tasks. By building upon these foundational works, our research aims to contribute novel methodologies and insights to enhance the accuracy and efficiency of COVID-19 diagnosis and pneumonia assessment in medical imaging.

\subsection{VGG-16}
The VGG-16 model, a CNN architecture, was introduced by Simonyan and Zisserman \cite{Simonyan2015} \cite{Sharma2023}. It is renowned for its simplicity and effectiveness in image classification tasks. The "16" in its name refers to the total number of layers in the network, including 13 convolutional layers and 3 fully connected layers. VGG-16 has a uniform architecture with small 3x3 convolutional filters, followed by max-pooling layers to downsample the spatial dimensions of the feature maps. This architecture helps in learning hierarchical features of increasing complexity from input images. Despite its effectiveness, VGG-16 is computationally expensive and memory-intensive, which limits its usage in resource-constrained environments. However, it serves as a foundational model in deep learning. It has paved the way for more advanced architectures, such as ResNet and Inception, while still being widely used as a benchmark in research and industry applications.

\subsection{VGG-19 Model}
The VGG-19 model, an extension of the VGG-16 architecture, was also developed by Simonyan and Zisserman \cite{Simonyan2015} \cite{Jain2020}. Like VGG-16, it is a CNN designed for image classification tasks. The "19" in its name denotes the total number of layers, which includes 16 convolutional layers and 3 fully connected layers. VGG-19 follows a similar architecture to VGG-16, with stacks of 3x3 convolutional layers and max-pooling layers to downsample the feature maps. However, VGG-19 includes more convolutional layers than VGG-16, allowing it to capture more intricate features from input images. This deeper architecture enables VGG-19 to learn more complex representations but also increases computational requirements. As with VGG-16, VGG-19 is widely used as a benchmark in deep learning for tasks such as image classification, object detection, and feature extraction. While its deep architecture contributes to its effectiveness in capturing detailed features, it also makes it computationally expensive, limiting its usage in resource-constrained environments.

\subsection{InceptionV3 Model}
The Inception v3 model, developed by Szegedy et al. \cite{Szegedy2016} \cite{Jain2020}, is a CNN architecture designed for image classification and object recognition tasks. It improves upon its predecessor, Inception v1, with enhancements to improve accuracy and efficiency. One of the key features of Inception v3 is its use of "inception modules," which perform parallel convolutions at different spatial scales and concatenate their outputs. This design allows the network to capture a wide range of features at multiple levels of abstraction, leading to improved performance. Additionally, Inception v3 incorporates techniques such as batch normalization, factorized convolutions, and aggressive regularization to enhance performance further and reduce overfitting. Inception v3 is known for its high accuracy on benchmark datasets such as ImageNet, where it achieved top results in image classification competitions. It has also been widely used in various applications such as image recognition, object detection, and image segmentation. Despite its effectiveness, Inception v3 is computationally intensive, requiring significant computational resources for training and inference.

\subsection{ResNet-50 Model}
The ResNet-50 model, introduced by He et al. \cite{He2016} \cite{Manickam2021}, is a CNN architecture that belongs to the ResNet (Residual Network) family. It addresses the problem of vanishing gradients in deep neural networks by introducing skip connections or "residual blocks." ResNet-50 consists of 50 layers and employs these residual blocks to enable the training of very deep neural networks. These blocks contain shortcut connections that allow gradients to flow directly through the network, mitigating the degradation problem commonly encountered in deep networks. This architecture facilitates the training of deeper models without suffering from diminishing performance. ResNet-50 has achieved notable success in various computer vision tasks, including image classification, object detection, and image segmentation. It has been widely adopted in both research and industry due to its effectiveness in learning highly complex features from images while maintaining relatively efficient computational requirements compared to other deep architectures. Additionally, pre-trained versions of ResNet-50 on large datasets such as ImageNet are often used as feature extractors or fine-tuned for specific tasks, leveraging the learned representations for transfer learning.

\subsection{CheXNet}
CheXNet, developed by Rajpurkar et al. \cite{Rajpurkar2017}, is a deep learning model that targets explicitly detecting pneumonia from chest X-rays. Utilizing DenseNet,a 121-layer CNN, CheXNet surpasses previous approaches and even outperforms radiologists regarding pneumonia detection accuracy. The model is trained on a large dataset of chest X-ray images labeled with up to 14 different thoracic pathologies, making it a robust solution for identifying pneumonia and other common thorax diseases. CheXNet's effectiveness has been proven through extensive testing, where it achieved a high level of precision and recall, demonstrating the significant potential of deep learning technologies in transforming diagnostic practices in the healthcare sector.

\subsection{U-Net Architecture}
The U-Net architecture, initially developed by Ronneberger et al. \cite{Ronneberger2015}, has become a cornerstone in medical image segmentation. Its unique design, characterized by a contracting path to capture context and a symmetric expanding path that enables precise localization, has proven particularly effective for segmenting lung tissue and identifying pathologies in chest X-rays and CT images. The U-Net's ability to operate with limited training data while still producing high-quality segmentations has made it especially valuable in medical applications where annotated images are scarce. This architecture has been adapted and extended in numerous studies, such as those by Çiçek et al. \cite{Cicek2016} and Alom et al. \cite{Alom2018}, who have introduced modifications to enhance its performance and adaptability to new challenges within medical imaging.

\section{Methodology}

\begin{figure}[!htbp]
    \centering
    \includegraphics[width=0.8\textwidth]{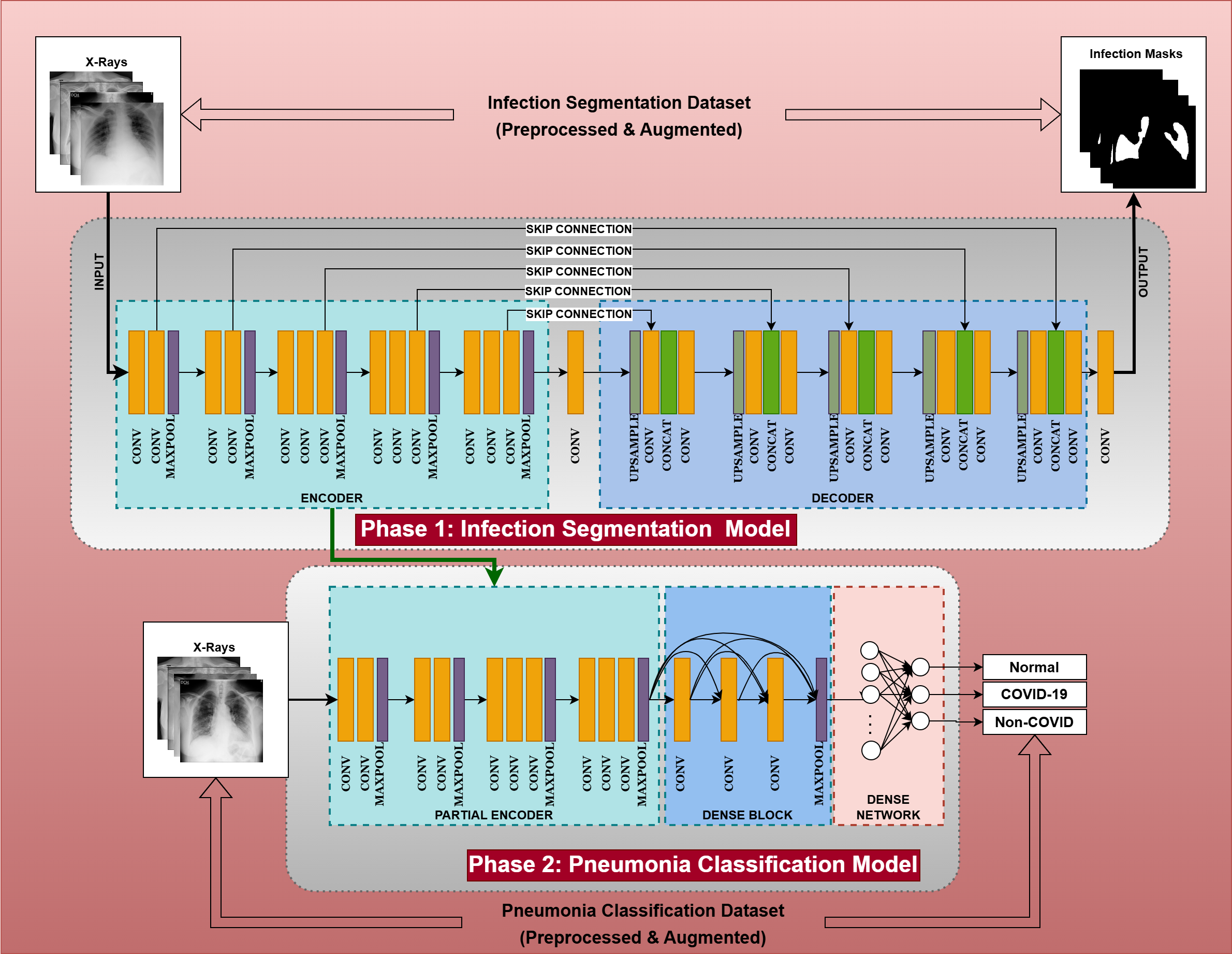}
    \caption{Training of Infection Segmentation and Pneumonia Classification Models.}
    \label{fig:Training}
\end{figure}

This study's methodology involves a two-phased approach as shown in Fig.~\ref{fig:Training} to developing and deploying deep learning models for the analysis of lung infections. These phases focus on incrementally training two interconnected models using transfer learning techniques.

\subsection{Phase 1: Infection Segmentation Model Training}
\begin{figure}[!htbp]
    \centering
    \includegraphics[width=0.8\textwidth]{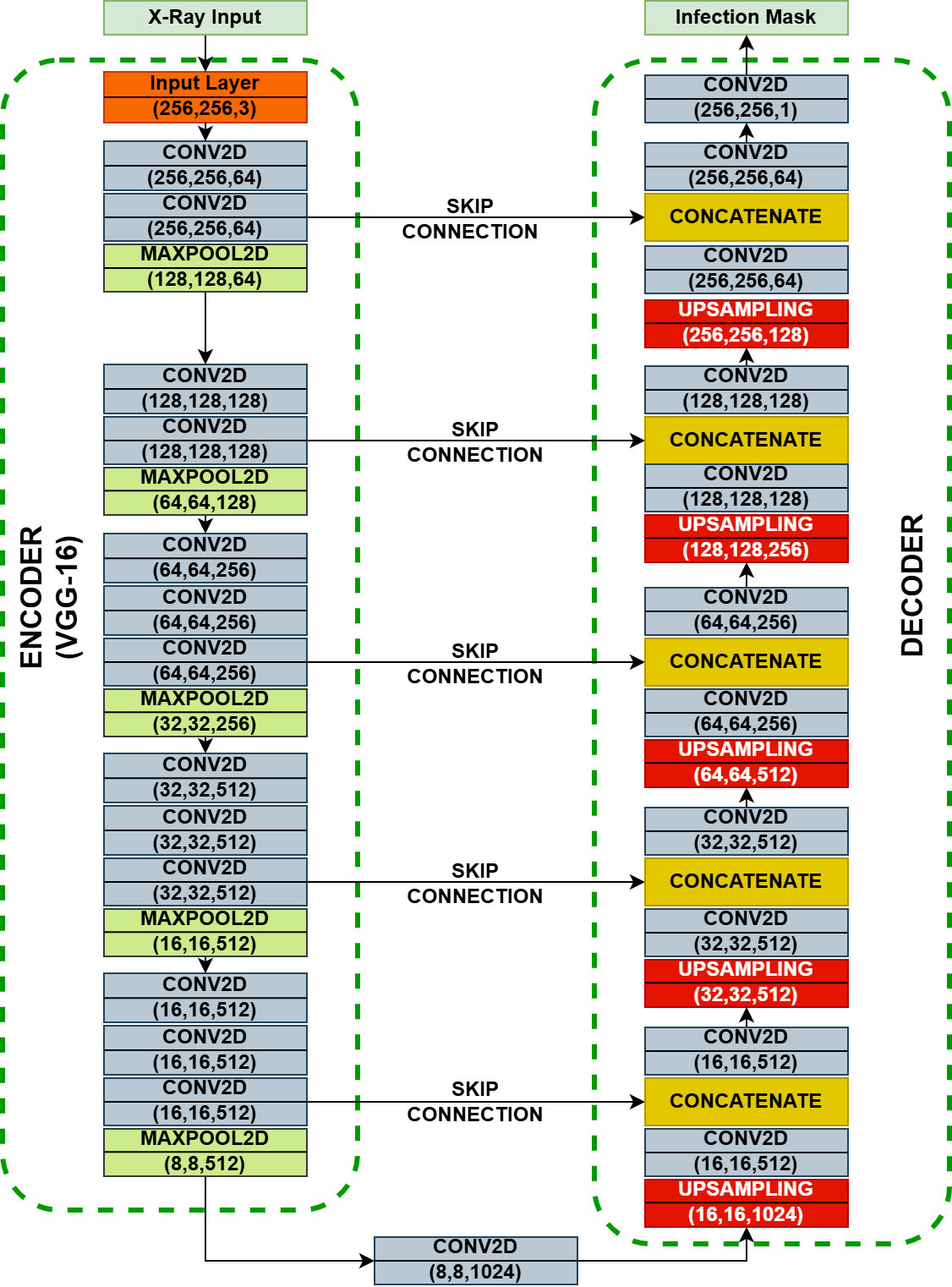}
    \caption{U-Net for Infection Segmentation.}
    \label{fig:UNET}
\end{figure}

The initial phase involves training an infection segmentation model based on the U-Net architecture, which incorporates a VGG16 convolutional network as its encoder, depicted in Fig.~\ref{fig:UNET}. The VGG16 blocks, renowned for their effectiveness in feature extraction from images, are initially frozen to leverage the pre-trained weights without altering them during the early stages of training. As training progresses, these layers are incrementally unfrozen stepwise, starting from the deeper layers and moving backward. This approach allows the model to refine its feature extraction capabilities more finely, adapting specifically to the nuances of lung infection imagery.

\subsection{Phase 2: Pneumonia Classification Model Development}
\begin{figure}[!htbp]
    \centering
    \includegraphics[width=0.95\textwidth]{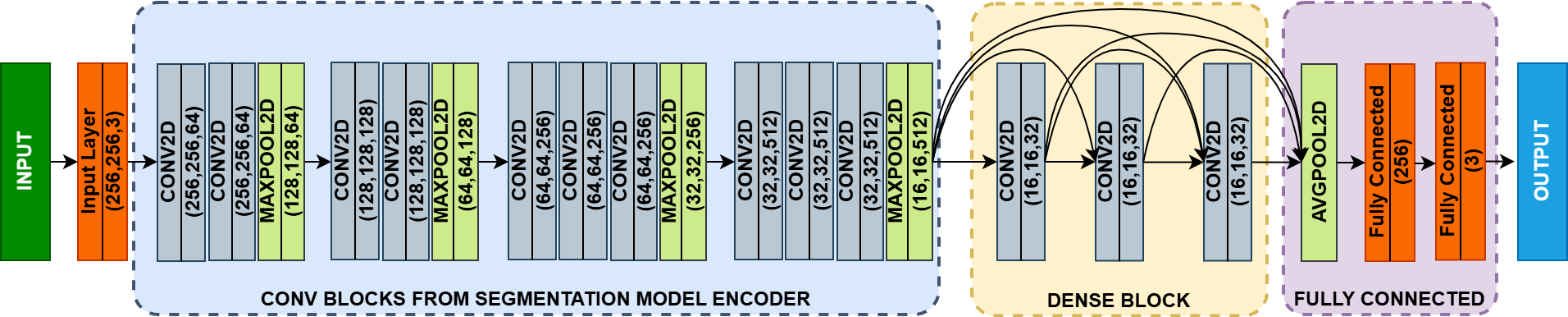}
    \caption{Pneumonia Classification Model.}
    \label{fig:Classifier}
\end{figure}

Utilizing the initial four blocks of the encoder part of the trained segmentation model followed by a dense block inspired by DenseNet121 (ChesXNet), a new classification model is developed to categorize X-ray images into Normal, Covid-19, and Non-Covid categories, as shown in Fig.~\ref{fig:Classifier}. This model extends the pre-trained encoder with additional dense blocks and fully connected layers, enhancing its ability to discern between the defined categories. Similar to the segmentation model, this classification model employs incremental unfreezing of its layers. The model is fine-tuned in stages to optimize performance, carefully adjusting to the specific characteristics of the classification task while maintaining the intricate patterns learned during the segmentation training.

\subsection{Prediction Workflow Using Trained Models}

\begin{figure}[!ht]
    \centering
    \includegraphics[width=0.9\textwidth]{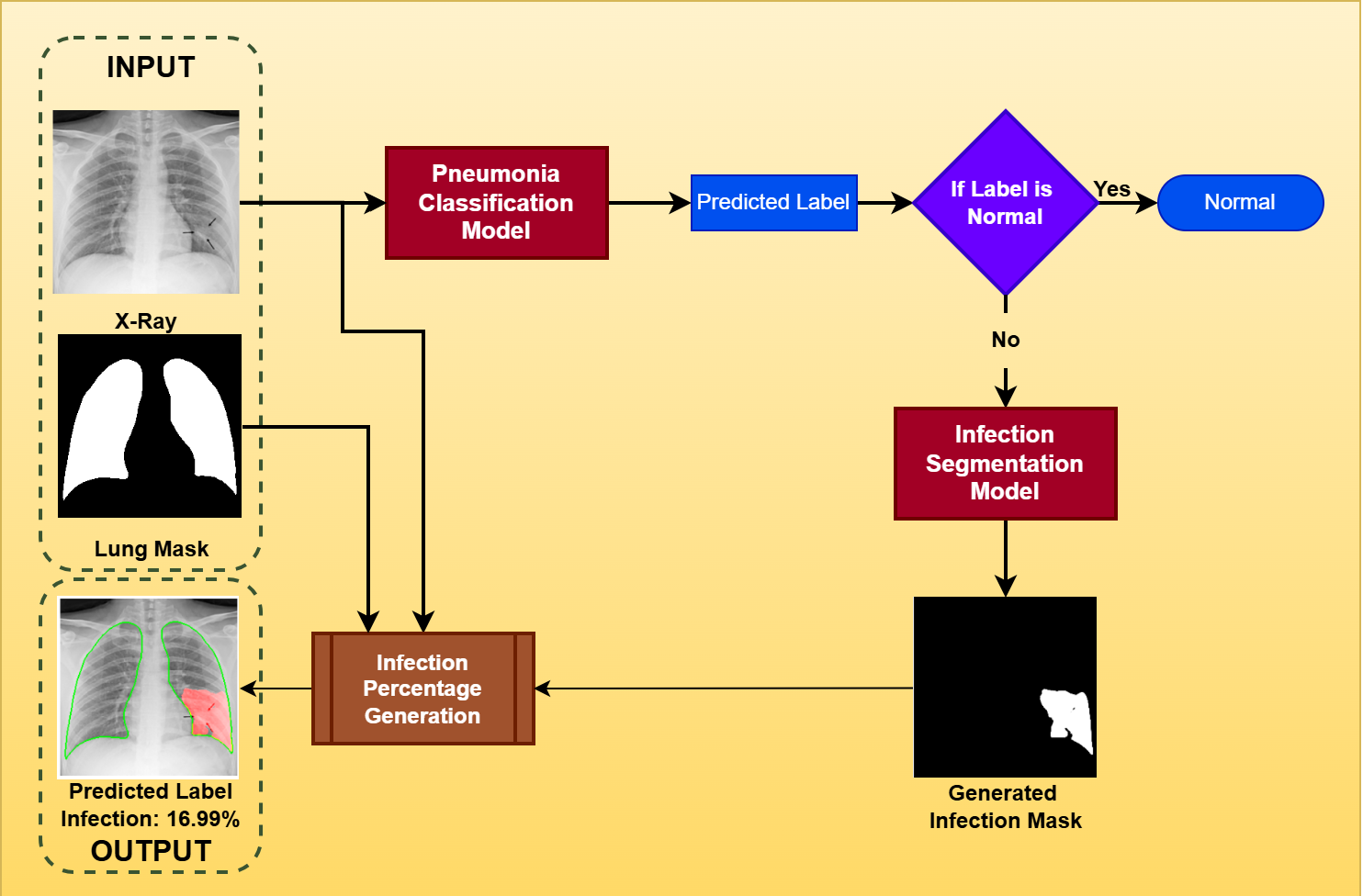}
    \caption{Workflow for making predictions using the trained models.}
    \label{fig:PredictionWorkflow}
\end{figure}

The prediction workflow is implemented using two trained deep learning models as described in previous sections. Figure \ref{fig:PredictionWorkflow} outlines the step-by-step process involved in making predictions for detecting and analyzing lung infections using these models.

Initially, an X-ray image is input into the pneumonia classification model, which categorizes it into one of three possible categories: Normal, COVID-19, or non-COVID. If the image is classified as Normal, the process terminates as no infection is detected. Images classified as COVID-19 or non-COVID are further processed using the infection segmentation model. This model generates an infection mask highlighting the infected regions within the lungs.

Subsequently, a lung mask and an infection mask are applied to the X-ray image. This combination allows for precise superimposition of the infected areas onto the lung regions of the X-ray, providing a clear visual representation for further analysis. The system calculates the percentage of the lung affected by the infection, which is crucial for determining the severity of the infection. The final output includes the classification result, the visual representation of the infection on the lung X-ray, and the quantified infection percentage, facilitating comprehensive evaluation and diagnosis.

\subsection{Performance Metrics}

Performance metrics are crucial for evaluating the effectiveness and accuracy of medical diagnostic models. This study utilized several metrics to assess the performance of both the infection segmentation and pneumonia classification models. The primary metrics included Accuracy, Precision, Recall, and the F1-Score.

\begin{itemize}
    \item \textbf{Accuracy:} This metric measures the overall correctness of the model across all classes. It is the ratio of correctly predicted observations to the total observations.
    \begin{equation}
    Accuracy = \frac{TP + TN}{TP + TN + FP + FN}
    \label{eq:accuracy}
    \end{equation}
    where \( TP \), \( TN \), \( FP \), and \( FN \) represent the true positives, true negatives, false positives, and false negatives, respectively.

    \item \textbf{Precision:} Also known as the positive predictive value, this metric assesses the model's ability to identify only relevant objects.
    \begin{equation}
    Precision = \frac{TP}{TP + FP}
    \label{eq:precision}
    \end{equation}

    \item \textbf{Recall:} Also known as sensitivity, this metric measures the model’s ability to identify all relevant cases.
    \begin{equation}
    Recall = \frac{TP}{TP + FN}
    \label{eq:recall}
    \end{equation}

    \item \textbf{F1-Score:} A harmonic mean of Precision and Recall, the F1-Score is crucial for situations where it is important to balance precision and recall, particularly in medical diagnostics, where the costs of false positives and false negatives are significant.
    \begin{equation}
    F1 = 2 \cdot \frac{Precision \cdot Recall}{Precision + Recall}
    \label{eq:f1_score}
    \end{equation}
\end{itemize}

For the segmentation model, the Dice Coefficient loss and Intersection over Union (IoU) were additionally used to quantify the accuracy of the infection masks generated by the model compared to ground truth annotations. These spatial overlap indices are critical for assessing how well the predicted infection regions align with actual infected areas on the lungs, reflecting the model's practical utility in clinical settings.

\begin{itemize}
    \item \textbf{Dice Coefficient:} This metric is used to gauge the similarity between the predicted segmentation and the ground truth.
    \begin{equation}
    Dice = \frac{2 \cdot TP}{2 \cdot TP + FP + FN}
    \label{eq:dice}
    \end{equation}

    \item \textbf{Intersection over Union (IoU):} Also known as the Jaccard index, this measures the overlap between predicted areas and ground truth.
    \begin{equation}
    IoU = \frac{TP}{TP + FP + FN}
    \label{eq:iou}
    \end{equation}
\end{itemize}

\subsection{Dataset Collection and Handling}
Our study employed two distinct datasets to address both infection segmentation and pneumonia classification challenges effectively:

\begin{enumerate}
    \item \textbf{Infection Segmentation Dataset:} The QaTa-COV19 dataset \cite{Degerli2021} is utilized for training our infection segmentation model. This dataset is significant as it comprises the largest collection of 119,316 chest X-ray images, including 2,951 COVID-19 samples. Each image is annotated with ground-truth segmentation masks of COVID-19 infected regions, carefully labeled using a novel collaborative human-machine approach. The dataset is particularly valuable for its capability to support the localization, severity grading, and detection of COVID-19 from chest X-ray images. Detailed experiments demonstrated that segmentation networks trained on this dataset could localize COVID-19 infection with high accuracy, significantly outperforming previous methods.

    \item \textbf{Pneumonia Classification Dataset:} The COVID-19 Radiography Database \cite{Chowdhury2020}, winner of the COVID-19 Dataset Award by the Kaggle Community, serves as the basis for our pneumonia classification model. This dataset includes a diverse collection of chest X-ray images categorized into COVID-19-positive cases and normal and viral pneumonia conditions. The database has been progressively updated and includes 3,616 COVID-19-positive cases, 10,192 normal images, 6,012 lung opacity (non-COVID lung infection), 1,345 viral pneumonia images, and corresponding lung masks.
\end{enumerate}

\textbf{Data Preprocessing:}
Before being input into the convolutional neural networks, both datasets undergo a series of preprocessing steps designed to optimize the learning efficiency and effectiveness of the models:

\begin{itemize}
    \item \textbf{Image Resizing:} All chest X-ray images are resized to a uniform dimension of 256x256 pixels. This resizing ensures that all samples' input sizes are consistent, facilitating the learning process without imposing excessive computational demands.
    \item \textbf{Normalization:} Pixel values of each image are normalized to have a mean of zero and a standard deviation of one. This step is crucial for converging the neural network quickly during training. Normalization helps in reducing disparities in image brightness and contrast, which can vary significantly across different X-ray images.
    \item \textbf{Augmentation:} To further enhance our models' robustness against overfitting and improve their ability to generalize across unseen data, we employ image augmentation techniques such as random rotations, zoom, and horizontal flipping during the training phase.
\end{itemize}

\subsection{Computational Environment}
The computational infrastructure for training and developing our models comprises high-end hardware and software components configured to handle extensive computational demands efficiently:

\begin{itemize}
    \item \textbf{CPU:} AMD Ryzen 7 5800X, renowned for its robust performance in executing complex parallel computations required for deep learning.
    \item \textbf{RAM:} 64 GB, facilitating efficient data handling and manipulation during intensive model training sessions.
    \item \textbf{GPU:} NVIDIA RTX 3090, equipped with 24 GB of VRAM and 10,496 CUDA cores, greatly accelerates the training process and allows for handling larger batches and more complex models seamlessly.
    \item \textbf{Operating System:} The system runs on a Linux platform, chosen for its stability and superior support for scientific computing and deep learning tasks.
\end{itemize}

This computational setup ensures our models are trained in an optimized environment, enabling faster iterations and robust performance evaluations.

\section{Results and Analysis}

This section outlines the outcomes of the deep learning models applied to lung infection detection tasks, focusing on infection segmentation and pneumonia classification. We present a detailed analysis of the model performances through various metrics and visual outputs.

\subsection{Infection Segmentation}
\begin{figure}[!htbp]
    \centering
    \includegraphics[width=0.9\textwidth]{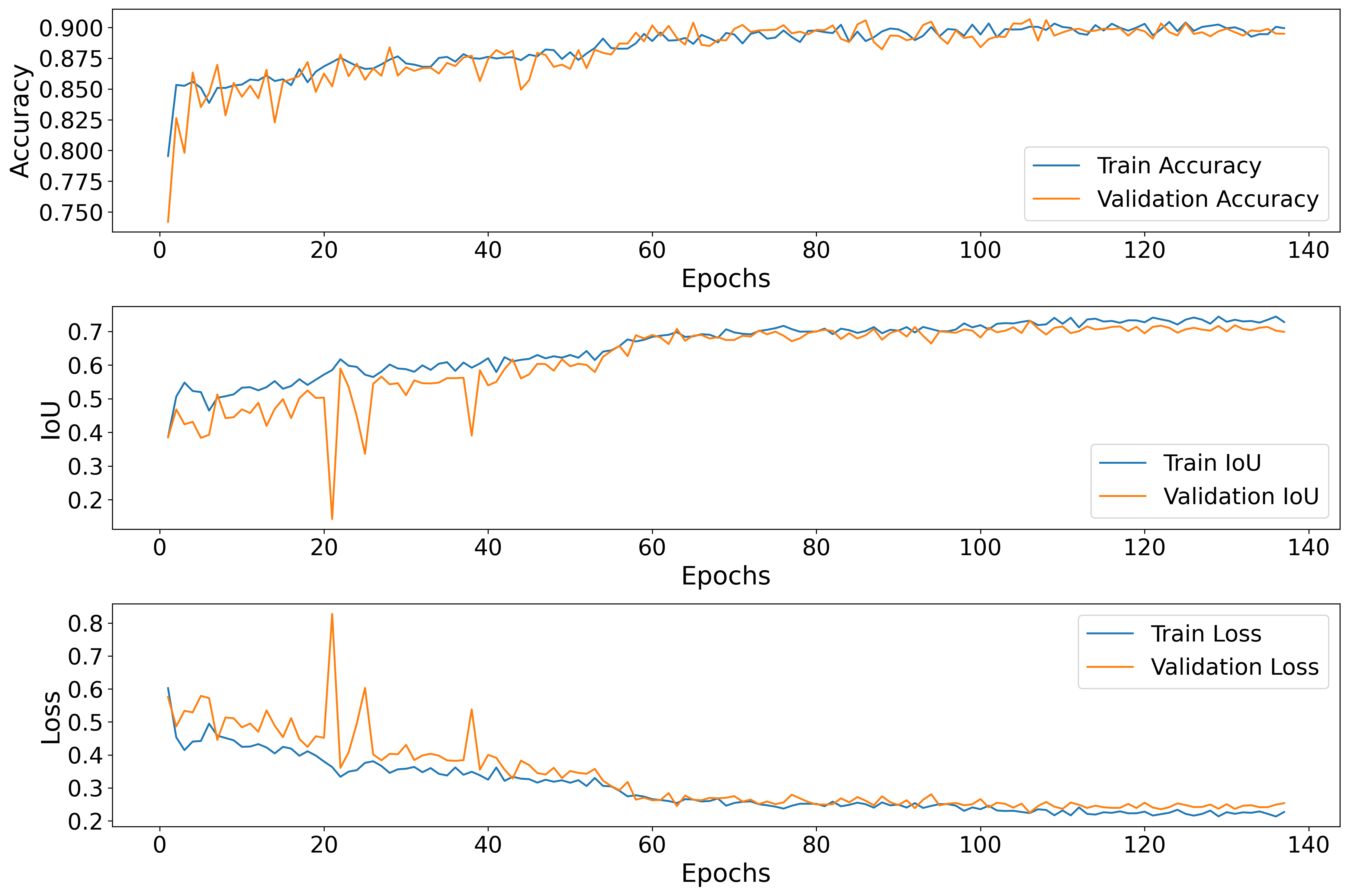}
    \caption{Training performance of the infection segmentation model.}
    \label{fig:seg_graph}
\end{figure}

The infection segmentation model was trained on the QaTa-COV19 dataset, utilizing a U-Net architecture with VGG16 as the encoder. Figure \ref{fig:seg_graph} illustrates the training process over 120 epochs, showing stability in the model's accuracy and Intersection over Union (IoU) metrics, alongside a decrease in loss, indicating effective learning and generalization capabilities.

\begin{figure}[!htbp]
    \centering
    \includegraphics[width=0.75\textwidth]{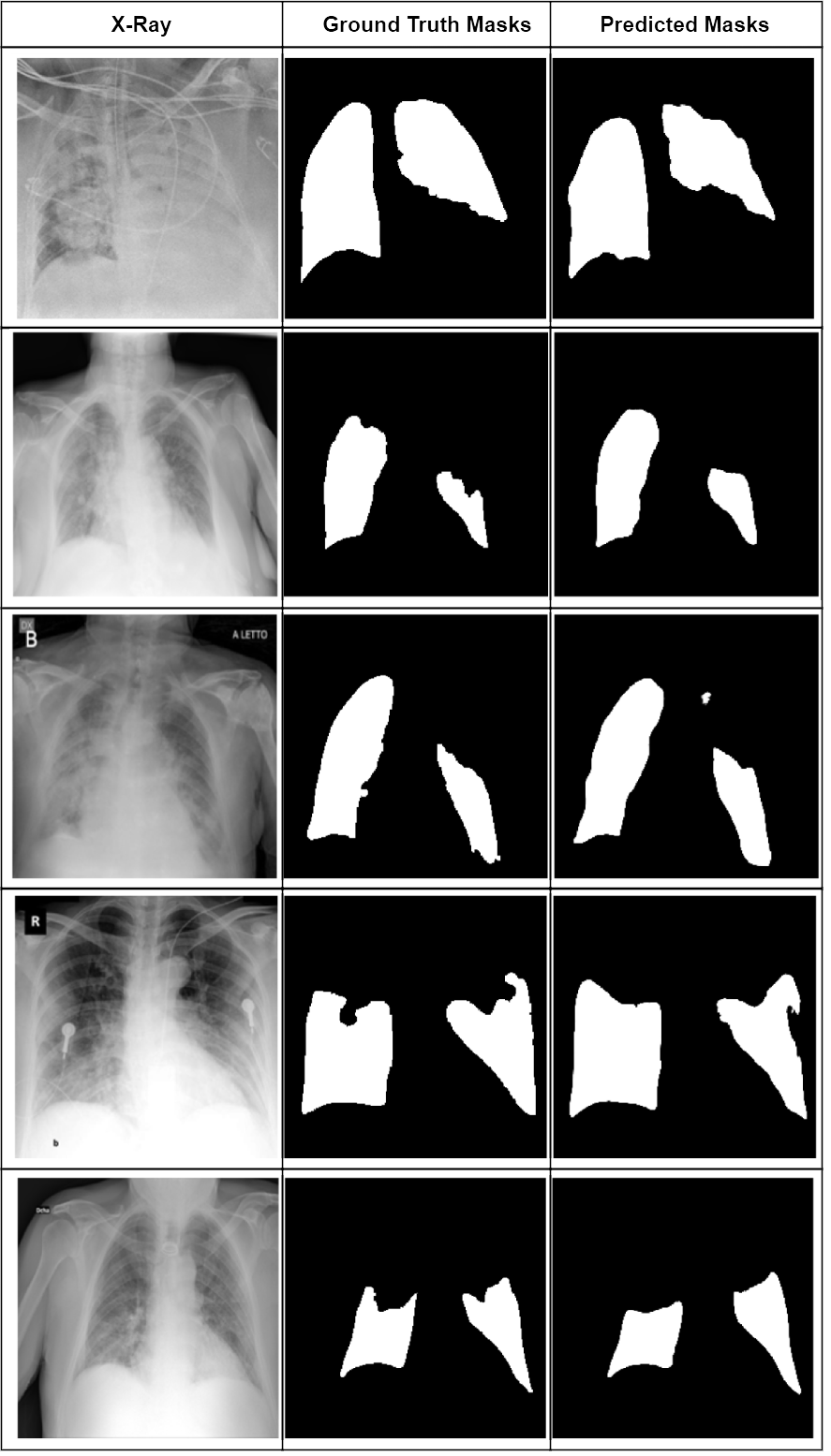}
    \caption{Comparison of original X-ray images, ground truth masks, and model predictions for infection segmentation.}
    \label{fig:xray_comparisons}
\end{figure}

\quad Further demonstrating the model's effectiveness, Figure \ref{fig:xray_comparisons} provides a side-by-side comparison of five X-ray images from the test set alongside their respective ground truth masks and the predictions generated by the model. These results highlight the precision of the segmentation model in identifying and delineating infected areas, with the predicted masks closely mirroring the expert-labeled ground truths.

\subsection{Pneumonia Classification Performance Analysis}

The pneumonia classification phase evaluated several convolutional neural network architectures to determine the most effective model for classifying chest X-rays into Normal, COVID-19, and non-COVID categories. A comprehensive analysis of accuracy, F1-score, and the number of parameters was conducted to assess each model's performance. These metrics are summarized in Table \ref{tab:model_performance}. Our proposed model demonstrated superior performance, particularly notable for its balance between high accuracy and computational efficiency, characterized by significantly fewer parameters than the other reviewed architectures. This makes it an attractive choice for practical deployment, where computational resources may be constrained.

\begin{table}[H]
\centering
\setlength{\tabcolsep}{4pt} 
\begin{tabular}{lccccccc}
\toprule
\textbf{Model} & \multicolumn{3}{c}{\textbf{Accuracy (\%)}} & \multicolumn{3}{c}{\textbf{F1-Score (\%)}} & \textbf{Parameters} \\
\cmidrule(lr){2-4} \cmidrule(lr){5-7}
 & Training & Validation & Difference & Training & Validation & Difference & \\
\midrule
VGG16           & 99.66 & 97.95 & 1.71 & 99.67 & 97.98 & 1.69 & $\sim$165M \\
VGG19           & 99.32 & 97.89 & 1.43 & 99.31 & 97.85 & 1.46 & $\sim$171M \\
ResNet50        & 97.74 & 91.39 & 6.35 & 97.72 & 91.1  & 6.62 & $\sim$292M \\
InceptionV3     & 99.78 & 97.95 & 1.83 & 99.78 & 97.86 & 1.92 & $\sim$172M \\
ChesXNet & 99.68 & 98.22 & 1.46 & 99.68 & 98.21 & 1.47 & $\sim$141M \\
Proposed        & 99.46 & 98.71 & 0.75 & 99.46 & 98.71 & 0.75 & $\sim$8M \\
\bottomrule
\end{tabular}
\caption{Comparison of CNN architectures for pneumonia classification.}
\label{tab:model_performance}
\end{table}
\subsubsection{VGG16 Model Performance}
\begin{figure}[!htbp]
    \centering
    \includegraphics[width=0.8\textwidth]{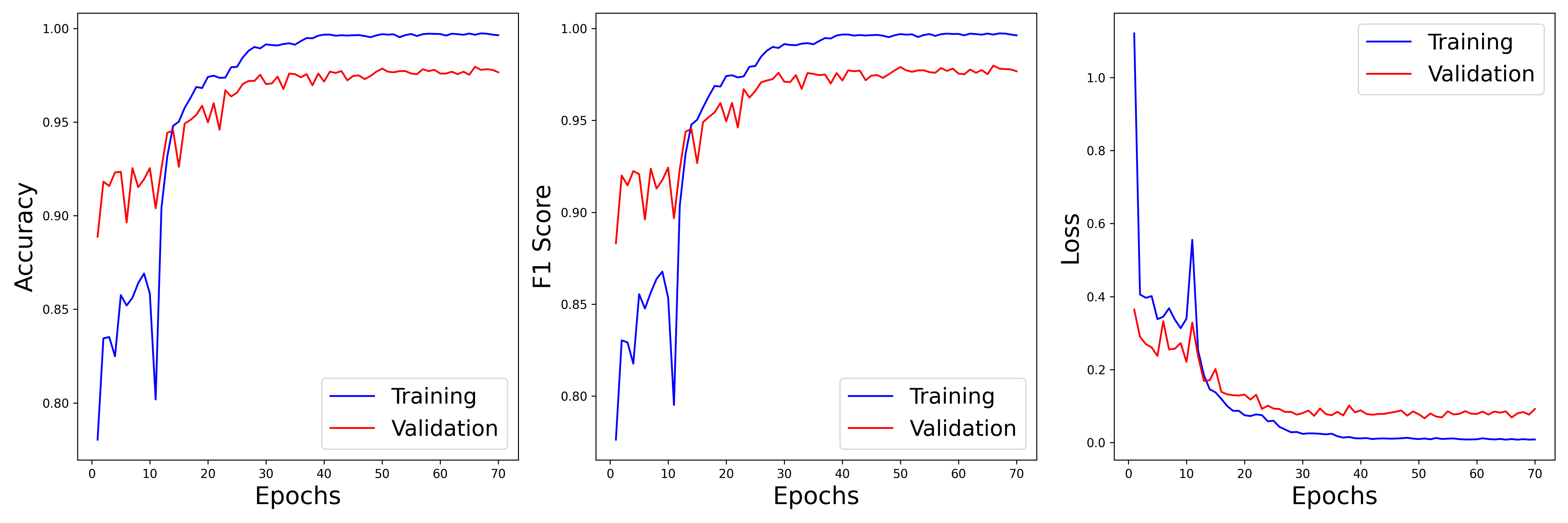}
    \caption{Training and validation performance of the VGG16 model.}
    \label{fig:VGG16}
\end{figure}

The VGG16 model showed robust performance with high accuracy and F1 scores. However, variations between training and validation metrics indicated possible overfitting issues, particularly in the later stages of training as depicted in Figure \ref{fig:VGG16}.

\subsubsection{VGG19, InceptionV3, and ResNet50 Models}
\begin{figure}[!htbp]
    \centering
    \includegraphics[width=0.8\textwidth]{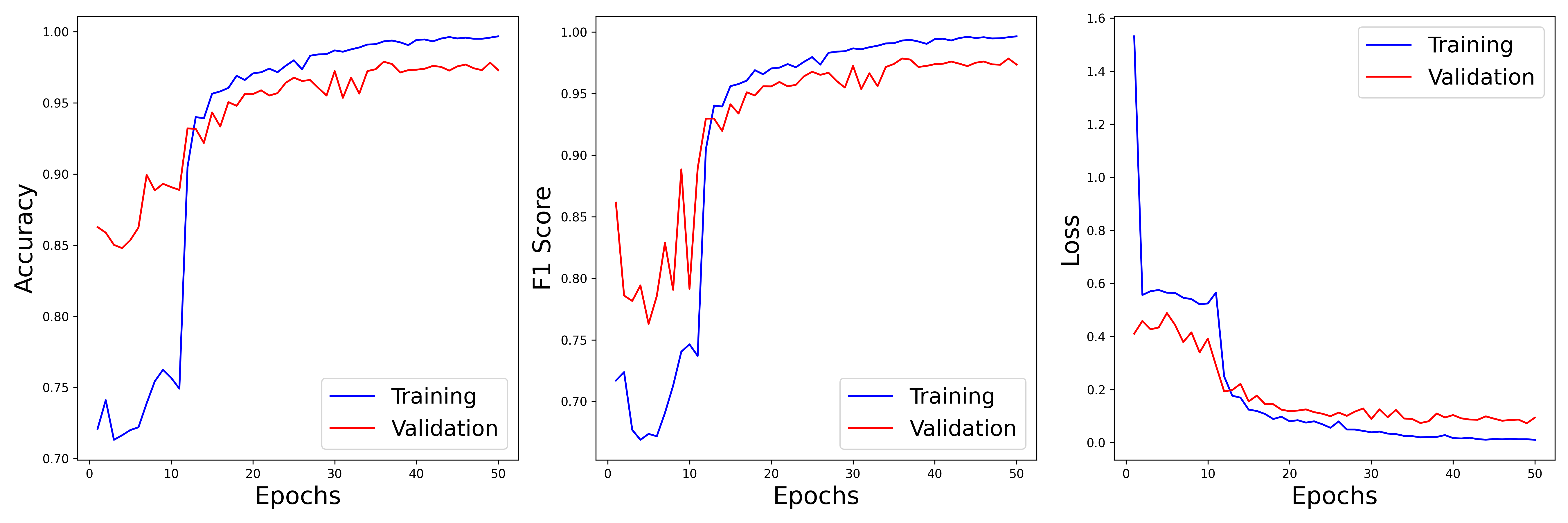}
    \caption{Training and validation performance of the VGG19 model.}
    \label{fig:VGG19}
\end{figure}

\begin{figure}[!htbp]
    \centering
    \includegraphics[width=0.8\textwidth]{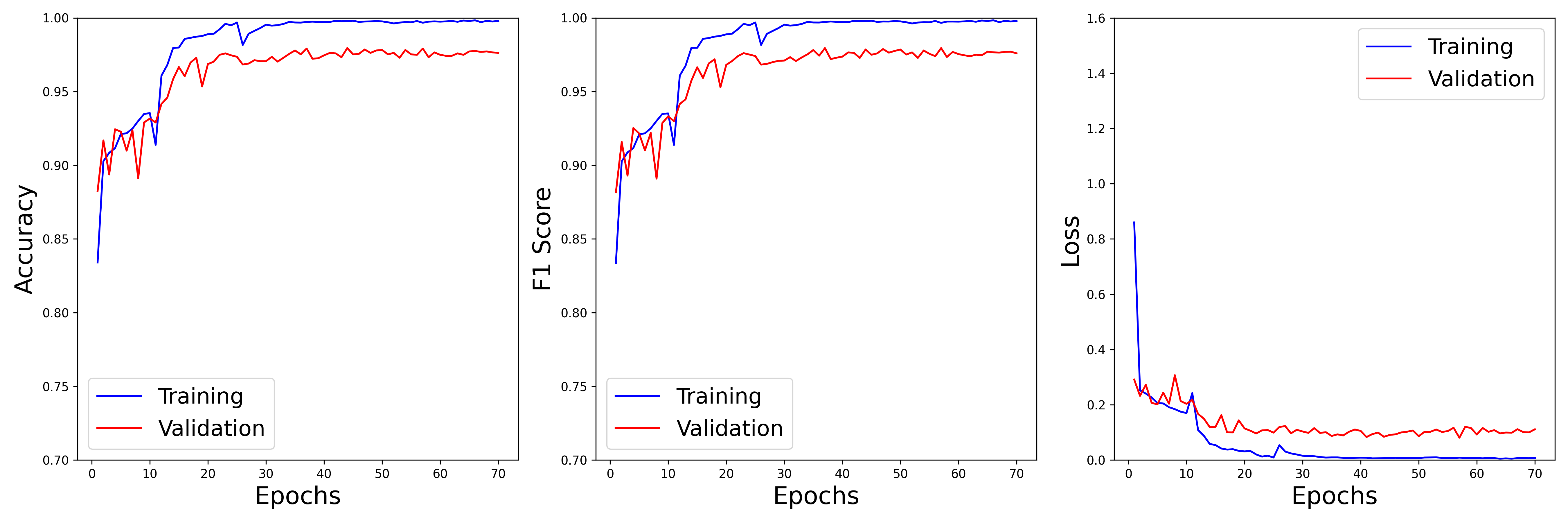}
    \caption{Training and validation performance of the InceptionV3 model.}
    \label{fig:Inceptionv3}
\end{figure}

\begin{figure}[!htbp]
    \centering
    \includegraphics[width=0.8\textwidth]{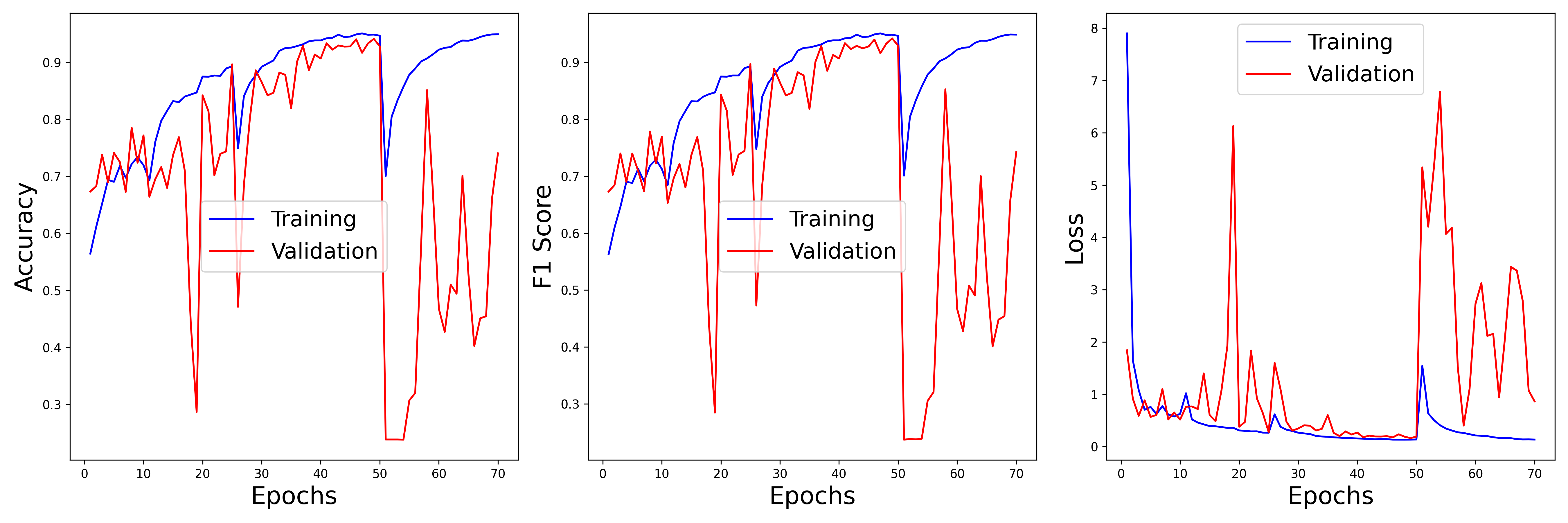}
    \caption{Training and validation performance of the ResNet50 model.}
    \label{fig:Resnet50}
\end{figure}

Subsequent experiments with VGG19, InceptionV3, and ResNet50 exhibited similar initial performance gains but with varying training-validation alignment levels. These results are illustrated in Figures \ref{fig:VGG19}, \ref{fig:Inceptionv3}, and \ref{fig:Resnet50} respectively.

\subsubsection{ChesXNet (DenseNet121) Adaptation}
\begin{figure}[!htbp]
    \centering
    \includegraphics[width=0.8\textwidth]{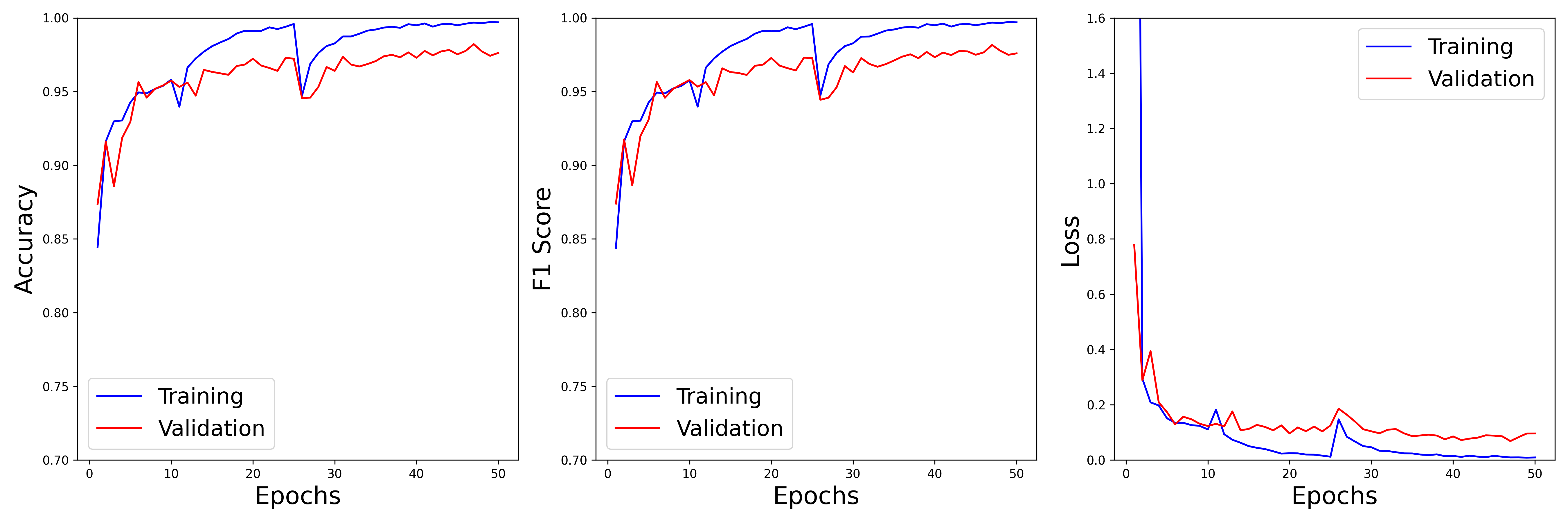}
    \caption{Training and validation performance of the ChesXNet (DenseNet121) model.}
    \label{fig:ChesXNetDenseNet121}
\end{figure}

The adaptation of ChesXNet, based on the DenseNet121 architecture, demonstrated a stable convergence between training and validation scores, proving its efficacy for X-ray image analysis. This stability is reflected in Figure \ref{fig:ChesXNetDenseNet121}.

\subsubsection{Optimized Model Combination}
\begin{figure}[!htbp]
    \centering
    \includegraphics[width=0.8\textwidth]{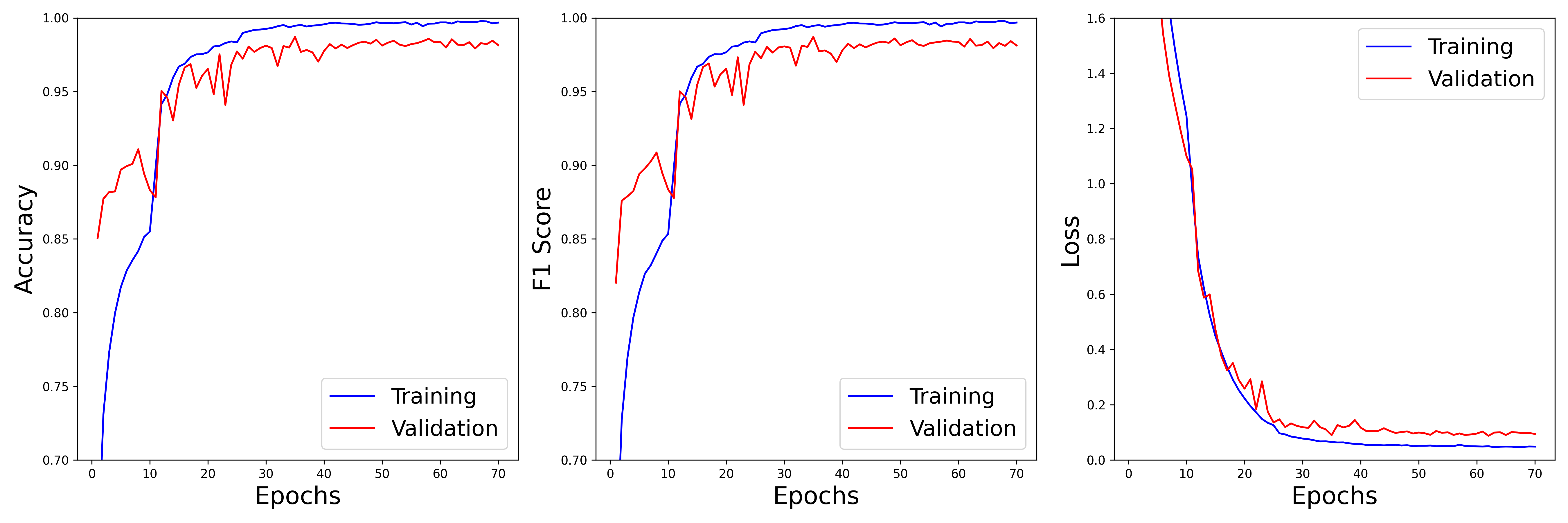}
    \caption{Enhanced validation performance using combined VGG16 and DenseNet blocks.}
    \label{fig:VGG16UNETFullVal}
\end{figure}

Integrating the four VGG16 convolutional blocks with a dense block resulted in significant validation performance improvements, ensuring high accuracy and better generalization capabilities, as shown in Figure \ref{fig:VGG16UNETFullVal}.

\subsection{System Operation and Final Output}

The final operational phase of our lung infection detection system capitalizes on the trained models to analyze chest X-ray images and highlight potential infection regions. This process is pivotal for aiding clinicians in the diagnosis of COVID-19 and non-COVID infections and differentiating these from normal cases.

\begin{figure}[ht]
    \centering
    \includegraphics[width=\textwidth]{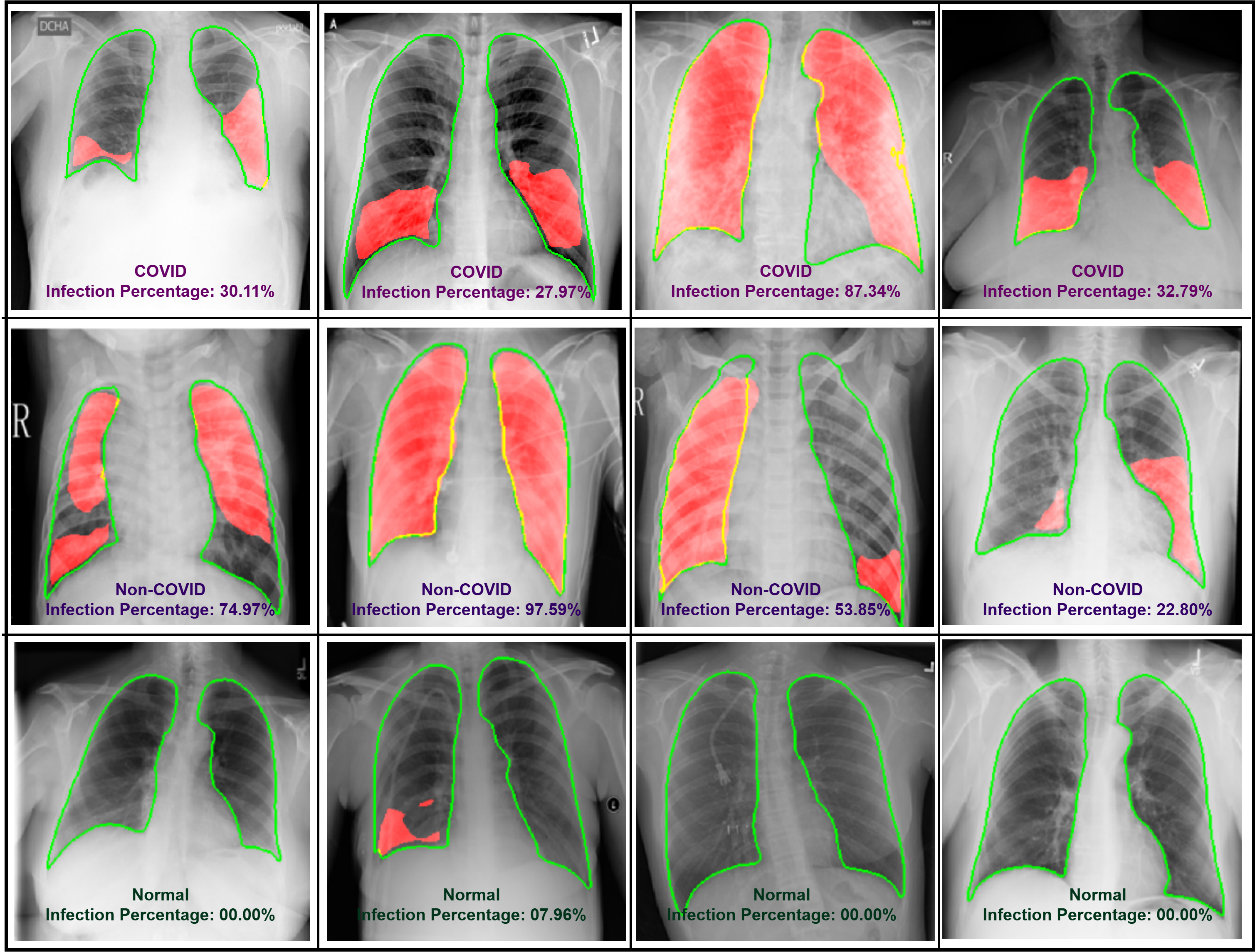}
    \caption{Visual output of the lung infection detection system showing segmented infection regions.}
    \label{fig:FinalResults}
\end{figure}

Figure \ref{fig:FinalResults} showcases a series of X-ray images processed through our system. The system accurately delineates lung boundaries for each image and identifies regions showing signs of infection. The infection segmentation model computes the percentage of the affected area, providing quantitative data that supports medical professionals in assessing the severity of the infection.

\begin{itemize}
    \item \textbf{COVID and Non-COVID Cases:} The images in the first two rows demonstrate cases with significant infection percentages, highlighted in red within the lung contours marked in green. These visualizations differentiate between COVID and non-COVID pneumonia, reflecting the model's capability to detect, localize, and quantify infection extents.
	\item \textbf{Normal Cases:} The last row presents normal cases, where the infection percentage is zero or close to zero. These examples are included for reference only to demonstrate the model's precision in recognizing healthy lung conditions without falsely detecting infections. The system does not generate infection masks for normal cases, ensuring efficient and focused diagnostic processing.
\end{itemize}

The system's ability to provide these detailed visual outputs enhances its utility in clinical environments where rapid and accurate assessments are crucial. Such detailed imagery and data facilitate a more informed decision-making process in patient care and treatment strategies.

\section{Model Explainability}

Model explainability is crucial in medical applications to validate the reliability and transparency of the predictive models used. This section discusses the application of Gradient-weighted Class Activation Mapping (Grad-CAM), which provides visual explanations for the decisions made by our convolutional neural network models.

\subsection{Grad-CAM Visualization}

\begin{figure}[ht]
    \centering
    \includegraphics[width=0.8\textwidth]{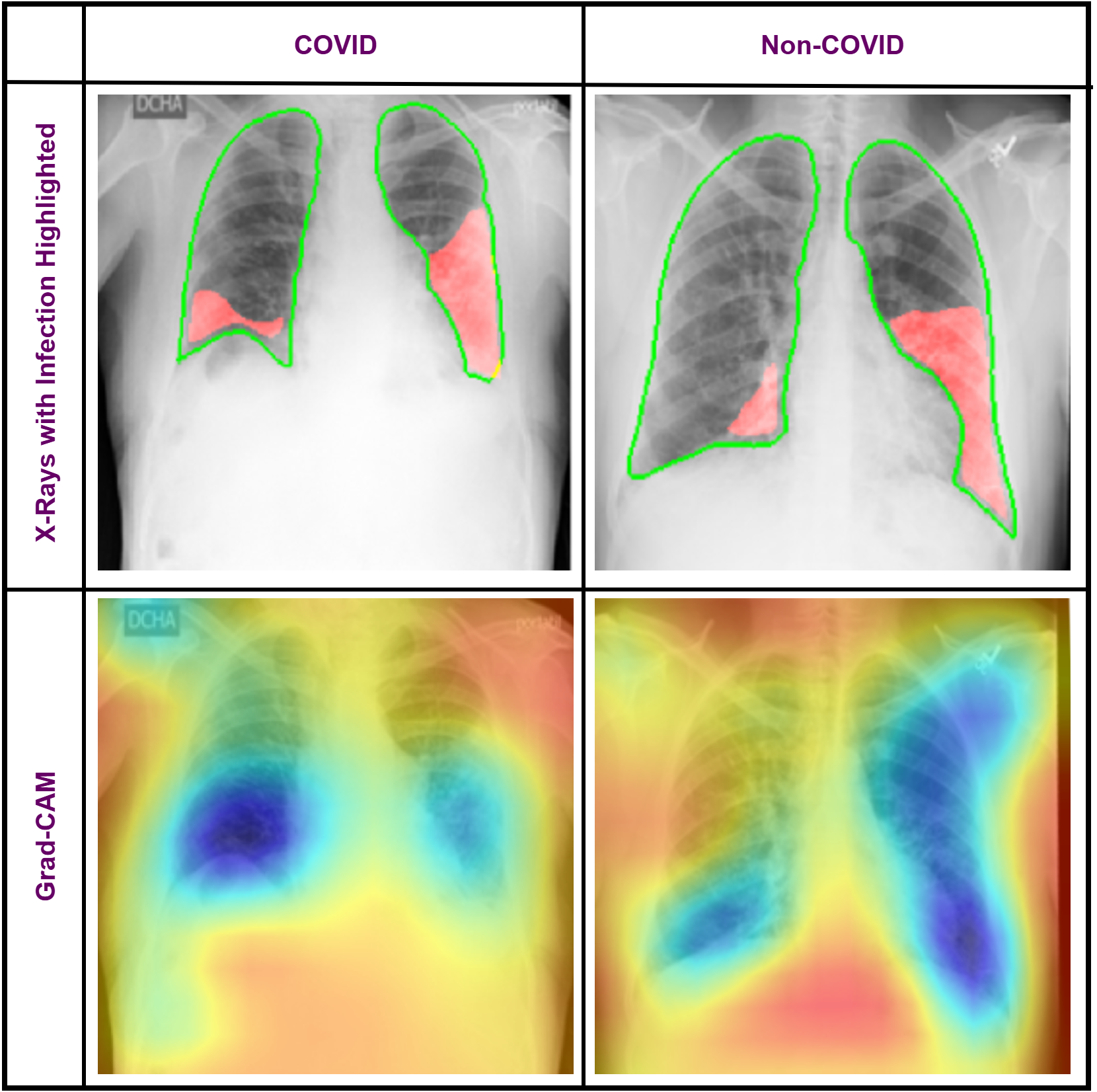}
    \caption{Grad-CAM heatmaps and corresponding X-rays highlighting infection regions for COVID and non-COVID cases.}
    \label{fig:GradCAM}
\end{figure}

Grad-CAM utilizes the gradients of any target concept, flowing into the final convolutional layer to produce a coarse localization map highlighting the important regions in the image for predicting the concept. Figure \ref{fig:GradCAM} illustrates this by showing the original X-rays with infection regions highlighted, accompanied by their respective Grad-CAM heatmaps.

The images in the top row are X-rays from COVID and non-COVID pneumonia cases, and the detected infection areas are highlighted. These areas are contoured to show the extent of the infections. The bottom row presents the Grad-CAM heatmaps, which display the regions significantly influencing the model's predictions. The heatmaps show clear, vibrant colors in the regions deemed crucial by the network, thus providing insight into the model's focus and basis for its classification:

\begin{itemize}
    \item \textbf{COVID Cases:} The heatmap shows intense activity over the infection areas, confirming that the model's attention is correctly focused on the pathological features associated with COVID-19.
    \item \textbf{Non-COVID Cases:} Similar to COVID cases, the model focuses on abnormal regions that signify non-COVID pneumonia, validating its ability to differentiate between various types of infections.
\end{itemize}

\section*{Conclusion}
\label{sec:conclusion}
In this study, we have demonstrated the effectiveness of advanced deep-learning techniques in diagnosing and analyzing pneumonia from chest X-rays, with a specific focus on COVID-19-related infections. By leveraging convolutional neural networks for classification and U-Net architectures for segmentation, our models provide detailed insights into infection patterns, substantially aiding in accurate medical diagnosis and treatment planning. Integrating these technologies not only enhances the accuracy of pneumonia detection but also offers significant promise in improving patient outcomes by facilitating timely and precise interventions.

\section*{Future Work}
\label{sec:futurework}
The promising results obtained in this study open several avenues for future research and development, which are outlined below:

\begin{itemize}
    \item \textbf{Multi-Modal Data Integration:} Future iterations could integrate additional data types such as patient demographics, historical medical records, and laboratory test results with the imaging data. This integration would potentially improve diagnostic accuracy by providing a holistic view of the patient's health.
    
    \item \textbf{Transfer and Federated Learning:} Implementing transfer learning could reduce the need for extensive local dataset collections by adapting pre-trained models to new tasks. Additionally, federated learning could be explored to build robust models across various institutions while ensuring privacy and data security, thus enhancing the model's applicability in diverse clinical environments.
    
    \item \textbf{Real-Time Diagnostic Tools:} Developing real-time diagnostic tools that leverage deep learning to provide immediate assessments of medical images would be crucial, especially for deployment in resource-limited environments where healthcare services are under pressure.
    
    \item \textbf{Telemedicine Enhancements:} Integrating these deep learning models into telemedicine platforms can facilitate remote diagnosis and monitoring, expanding the reach of high-quality healthcare services and reducing the risks associated with in-person consultations during pandemics.
    
    \item \textbf{Continuous Model Refinement:} Continuous improvement of these models is essential to adapt to evolving medical standards and emerging new data. This involves regular updates and validations to ensure accuracy and reliability in diverse clinical settings.
\end{itemize}

\section*{Declarations}

\textbf{Clinical Applicability and Limitations:} \\
This research involved the development and evaluation of machine learning models designed to assist in the diagnosis of pneumonia from chest X-ray images, utilizing well-established, publicly available datasets annotated by medical professionals, including the COVID-19 RADIOGRAPHY DATABASE and the QaTa-COV19 dataset. While the models show promising results, they are intended to support, not replace, professional medical diagnosis. They do not incorporate clinical patient data, crucial for comprehensive diagnosis in real-world settings. The models' predictions are probabilistic and should be used as an adjunct to traditional diagnostic methods under the supervision of qualified healthcare professionals, with diagnostic decisions made considering all clinical evaluations and diagnostic tools available.
\bibliographystyle{apalike}
\bibliography{Article.bib}

\end{document}